\documentclass[sigconf,screen,nonacm]{acmart}

\usepackage{algorithm}
\usepackage{algorithmic}
\usepackage{multirow}
\usepackage{enumitem}
\usepackage[dvipsnames]{xcolor}

\AtBeginDocument{%
  }

\renewcommand\footnotetextcopyrightpermission[1]{}

\begin{document}


\title[Creative4U]{Creative4U: MLLMs-based Advertising Creative Image Selector \\ with Comparative Reasoning}




\settopmatter{authorsperrow=5}

\author{Yukang Lin}
\authornote{Both authors contributed equally to this research.}
\affiliation{%
  \institution{Tsinghua University}
  \country{China}
}

\author{Xiao Zhang}
\authornotemark[1]
\affiliation{%
  \institution{Alibaba Group}
  \country{China}
}

\author{Xiang Zhang}
\authornotemark[1]
\affiliation{%
  \institution{Alibaba Group}
  \country{China}
}

\author{Shichang Jia}
\affiliation{%
  \institution{USTB}
  \country{China}
}

\author{Bowen Wan}
\affiliation{%
  \institution{USTB}
  \country{China}
}

\author{Chenghan Fu}
\affiliation{%
  \institution{Alibaba Group}
  \country{China}
}

\author{Xudong Ren}
\affiliation{%
  \institution{Alibaba Group}
  \country{China}
}

\author{Yueran Liu}
\affiliation{%
  \institution{Alibaba Group}
  \country{China}
}

\author{Wanxian Guan}
\affiliation{%
  \institution{Alibaba Group}
  \country{China}
}

\author{Pengji Wang}
\affiliation{%
  \institution{Alibaba Group}
  \country{China}
}

\author{Chuan Yu}
\authornote{Both authors are the corresponding authors.}
\affiliation{%
  \institution{Alibaba Group}
  \country{China}
}

\author{Baolin Liu}
\authornotemark[2]
\affiliation{%
  \institution{USTB}
  \country{China}
}


\renewcommand{\shortauthors}{Lin et al.}

\begin{abstract}
    Creative image in advertising is the heart and soul of e-commerce platform. An eye-catching creative image can enhance the shopping experience for users, boosting income for advertisers and advertising revenue for platforms. With the advent of AIGC technology, advertisers can produce large quantities of creative images at minimal cost. However, they struggle to assess the creative quality to select. Existing methods primarily focus on creative ranking, which fails to address the need for explainable creative selection.
    
    In this work, we propose the first paradigm for explainable creative assessment and selection. Powered by multimodal large language models (MLLMs), our approach integrates the assessment and selection of creative images into a natural language generation task. To facilitate this research, we construct \textbf{CreativePair}, the first comparative reasoning-induced creative dataset featuring 8K annotated image pairs, with each sample including a label indicating which image is superior. Additionally, we introduce \textbf{Creative4U} (pronounced \textit{Creative for You}), a MLLMs-based creative selector that takes into account users' interests. 
    To train Creative4U, we propose Group Relative Policy Optimization with Recall and Precision Reward, namely \textbf{GRPO-RPR}, which explicitly optimizes the recall and precision of the comparative reasoning, enabling the model to both accurately select creatives and provide high-quality explanations.
    Both offline and online experiments demonstrate the effectiveness of our approach. Our code and dataset will be made public to advance research and industrial applications.
\end{abstract}

\ccsdesc[500]{Computing methodologies~Computer vision}

\keywords{Advertising creative selection, Image quality assessment, Multimodal large language model}

\begin{teaserfigure}
    \centering
  \includegraphics[width=0.98\textwidth]{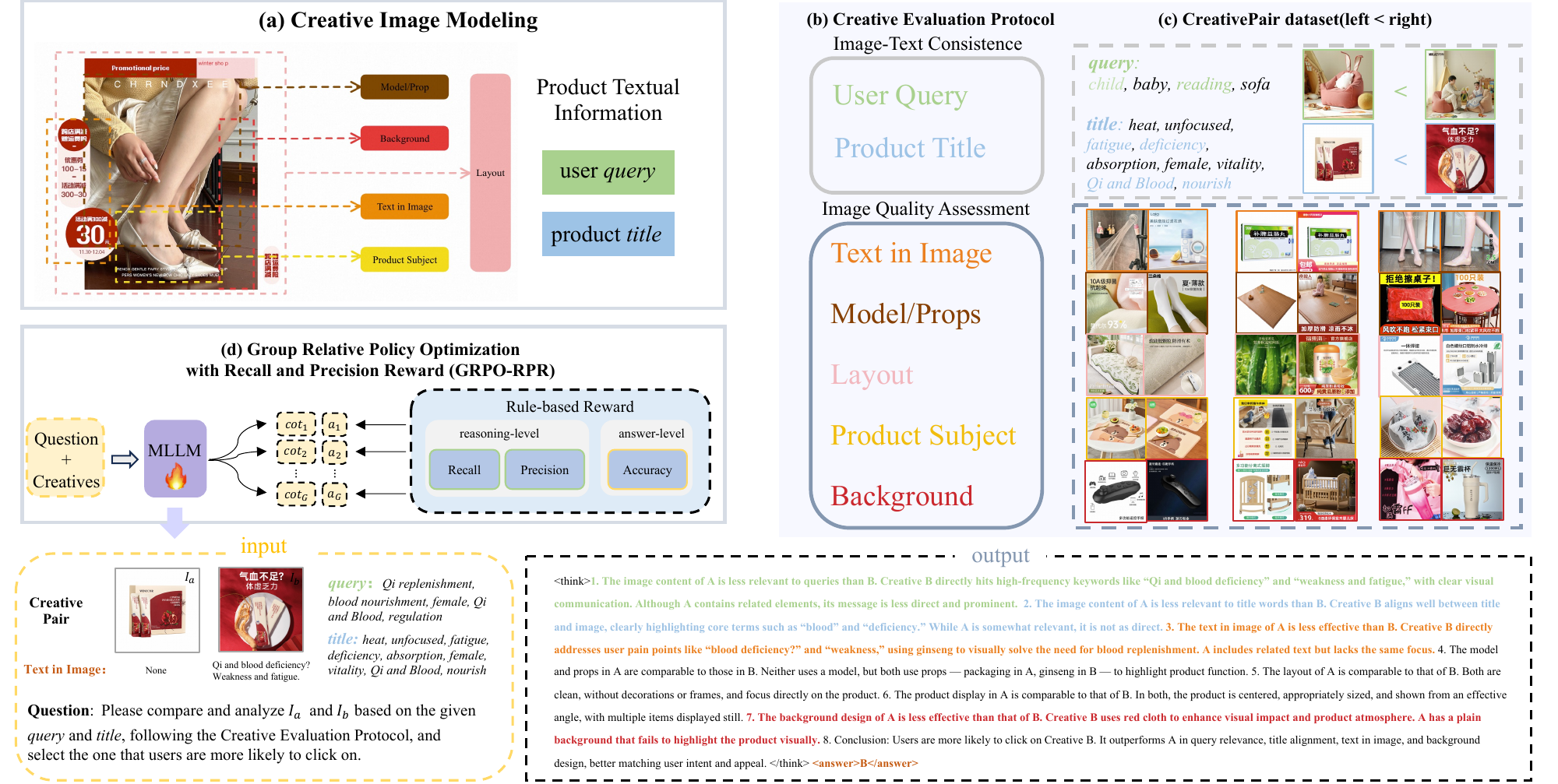}
  \caption{A Holistic Snapshot of Creative4U.  To systematically analyze creative images, we first perform a comprehensive modeling(a), based on which we design a structured Creative Evaluation Protocol(b) that facilitates the assessment and selection of creative images. To ensure explainable selection, we construct the CreativePair dataset(c), providing pair-wise samples for comparative analysis. Through the GRPO-RPR(d), Creative4U can accurately assess and select creative images.} 
  \label{fig:teaser}
\end{teaserfigure}



\maketitle

\section{Introduction}

In the digital era, online advertising has become a pivotal income stream for leading e-commerce platforms such as Amazon, Shopee and Taobao, fundamentally reshaping their revenue generation landscape. Creative images represent one of the most prevalent forms of online advertising, primarily due to their intuitiveness, effectiveness, and ease of understanding. 
To capture user attention and convey product messages, advertisers typically engage professional designers to create visually striking and semantically clear creative images. With the emergence of AI-generated content (AIGC) technologies, advertisers can now produce a substantial amount of visually diverse and semantically relevant creative images at minimal cost. Nevertheless, advertisers are confronted with a critical \textbf{dilemma}: the challenge of determining which specific images to display to their target users, as well as the lack of systematic guidelines for improving the quality of visual creatives.

\begin{figure}
    \centering
    \includegraphics[width=\linewidth]{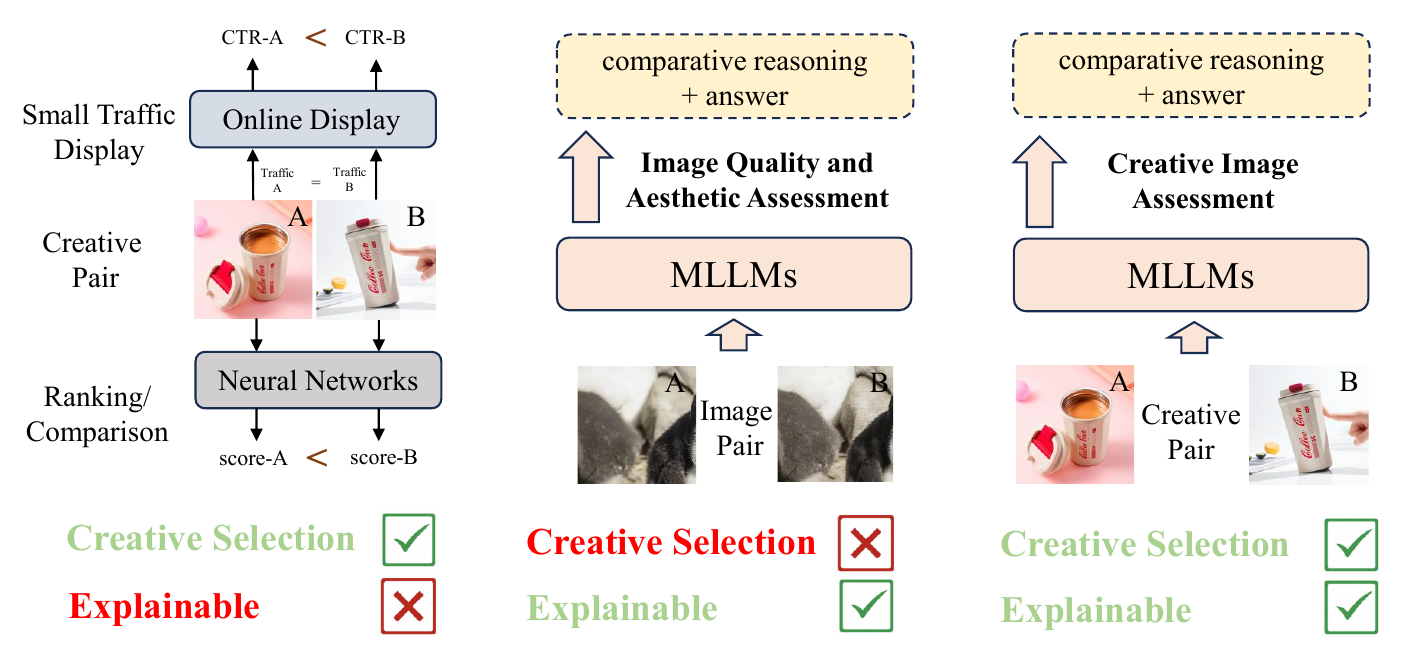}
    \caption{Comparison of three types of image selection.}
    \label{fig:intro}
\end{figure}

As shown in Fig.~\ref{fig:intro}, to evaluate the effectiveness of creative images of one product, advertising platforms often employ an online display strategy, in which multiple images are displayed to users with small and equal traffic allocation. This process allows the platform to obtain real user feedback in the form of click-through rates (CTR), which are then used to identify and prioritize high-performing creative images for future campaigns. Although accurate, this strategy becomes increasingly costly and inefficient as the number of creative images grows. In the academic community, many studies have achieved accurate selection of advertising creative images via ranking strategies. However, both the online display strategy and ranking strategies fail to provide an explanation for their selection decisions, making it difficult for designers or AIGC systems to receive meaningful feedback for improvement. Recently, an increasing number of researchers have leveraged MLLMs for image quality assessment and comparative selection, often accompanied by explicit reasoning processes. However, they predominantly focus on the evaluation of the quality of low-level aspects and aesthetics. As far as we know, \textbf{there is no framework for reasoning-induced creative image selection}. 

To address this gap, we establish a novel paradigm for efficient and explainable selection of high-performing advertising creative images. And we adopt a pair-wise strategy to design a comparative reasoning pipeline for selection. 
To this end, we construct the first comparative reasoning-induced creative dataset \textbf{CreativePair}, to facilitate explainable creative image selection. Specifically, we begin by conducting a comprehensive modeling of creative images, based on which we formulate a systematic \textit{Creative Evaluation Protocol} grounded in empirical insights. 
We then collect pairs of creative images and manually annotate them using the proposed protocol, and further employ the open-source model Qwen2.5-VL-72B-Instruct to generate fine-grained Chain-of-Thought comparative reasoning for each pair, resulting in the CreativePair dataset with 8K samples.

Building on \textbf{CreativePair} dataset, we propose \textbf{Creative4U}, a comparative reasoning-induced advertising creative image selector powered by the multimodal large language model. 
Creative4U, pronounced \textit{Creative for You}, is specifically designed to recommend advertiser creative images that are more likely to be clicked by target users. From another perspective, it can be viewed as a system that helps deliver content that aligns better with user preferences, thereby enhancing both advertising performance and user experience.
To improve the user alignment in the selection process, we incorporate high-frequency product-related queries (e.g., ``stylish black T-shirt'') entered by users into the search box. These queries reflect real user intent and provide valuable insights into their preferences for the product. 
Based on extensive exploration, we propose Group Relative Policy Optimization with Recall and Precision Reward (\textbf{GRPO-RPR}) to train Creative4U. On top of the default answer-level reward that supervises whether the selected creative is correct, GRPO-RPR introduces two reasoning-level rewards on the generated comparative explanations: a recall reward that encourages covering key factors affecting creative performance, and a precision reward that penalizes irrelevant or spurious arguments. By jointly optimizing these rewards, GRPO-RPR improves both selection accuracy and the quality of comparative explanations.

Through comprehensive qualitative and quantitative evaluation on the \textbf{CreativePair} test set, we demonstrate that \textbf{Creative4U} outperforms state-of-the-art methods in selecting high-performing creative images. Moreover, a robust online A/B test further validates its effectiveness in real-world deployment.

In summary, we present an applied, deployable solution for selecting high-quality advertising creatives from massive AI generated candidates.
Our contributions are as follows:
\begin{itemize}[leftmargin=*,noitemsep,nolistsep]
    \item \textbf{Creative Evaluation Protocol}, a structured protocol for advertising creatives that specifies comparison dimensions and rules for explainable selection.
    \item \textbf{CreativePair}, the first open-source dataset built under this protocol, comprising 8K image pairs with chain-of-thought comparative reasoning annotations.
    \item \textbf{Creative4U with GRPO-RPR}, a reasoning-aware MLLM-based creative selector trained with dimension-level recall and precision rewards, validated through offline benchmarks and online A/B tests on Taobao.
\end{itemize}

\section{Related Work}

\subsection{Advertising Creative Selection}
The selection of advertising creatives is a pivotal task for enhancing online advertising effectiveness and platform revenue, drawing significant attention from both academia and industry.
Meanwhile, AIGC technologies~\cite{seedance2026,rombach2022high,esser2024flux,kling2024,yang2024hunyuan3d,lin2024consistent123,lin2025mvportrait,lin2025interanimate} have substantially increased the supply of advertising materials across modalities, making creative selection increasingly difficult.
Foundational studies focused on quantifying the importance of a creative's visual appearance. Azimi et al.\cite{azimi2012impactvisualappearanceuser} conducted a seminal data-driven analysis, establishing a strong correlation between hand-crafted visual characteristics, such as color harmony and composition, and the click-through rate (CTR), thus proving that visual content inherently matters. Based on this, the research moved to address the critical cold start problem for new creatives. To mitigate the costs of online exploration, PEAC\cite{zhao-2019-look} is an offline pre-ranking framework that uses deep learning to evaluate creatives based on content before they are served. Concurrently, other approaches sought to bridge offline evaluation with online adaptation. For instance, VAM-HBM\cite{wang2021hybrid} proposed a hybrid bandit model that leverages visual priors for initial selection and then dynamically refines its strategy through online feedback, effectively balancing exploration and exploitation.
The advent of programmatic advertising introduced the challenge of combinatorial explosion from composited elements. To tackle this, AES\cite{chen2021efficientoptimalselectioncomposited} modeled element relationships as a tree structure, enabling efficient optimal selection via dynamic programming. In a different approach, CECS\cite{zhang2023crosselementcombinatorialselectionmultielement} framed it as a sequence selection problem, using an encoder-decoder architecture to capture inter-element dependencies.
More recent work has expanded to address system-level architecture and user-centric factors. Moriwaki et al.\cite{moriwaki2020fatigueawareadcreativeselection} developed a fatigue-aware model that quantifies a user's ad fatigue based on exposure history and creative similarity, allowing more dynamic and user-friendly selections. To address system efficiency, Yang et al.\cite{yang2023parallelrankingadscreatives} proposed a novel parallel ranking architecture to reduce latency, coupled with a joint offline optimization model to maintain effectiveness.
While previous studies tackled practical challenges in advertising creative selection, they lack a deep analysis of visual content within creatives. 
To address this limitation, we propose reasoning-induced creative image selection, leveraging MLLMs to generate coherent textual rationales that align with user preferences and enhance decision accuracy.

\subsection{Image Quality Assessment}

Image Quality Assessment (IQA) aims to create computational models that align with human perceptual judgments. Traditional methods relied on handcrafted features to measure specific distortions, but they typically lack the flexibility to assess the diverse impairments found in real-world images. The advent of deep learning\cite{lecun2015deep, rumelhart1986learning} led to data-driven models, primarily Convolutional Neural Networks\cite{krizhevsky2012imagenet,simonyan2014very,he2016deep}, which learn quality-aware features to regress a Mean Opinion Score (MOS)\cite{wang2004image,sheikh2006image,han2025ntire}. While effective, these models often act as black boxes, providing a score without interpretable reasoning.
Recently, MLLMs have introduced a new paradigm, shifting the focus of IQA towards more flexible, interpretable, and text-driven evaluations\cite{wu2024comprehensivestudymultimodallarge}. This evolution is marked by a transition from predicting single numerical scores to generating rich, descriptive assessments. Early works in this area, such as DepictQA\cite{you2024depictingscoresadvancingimage}, pioneered tasks requiring models to describe distortions in detail. This trend has progressed from single-image analysis to comparative evaluations involving multiple images, as explored by models like Compare2Score\cite{zhu2024adaptiveimagequalityassessment} and Co-Instruct\cite{wu2024towards}. The success of these advanced models is critically dependent on high-quality, large-scale instruction-tuning datasets. Projects like Q-Instruct\cite{wu2023qinstructimprovinglowlevelvisual} and DepictQA-Wild\cite{you2024descriptiveimagequalityassessment} use sophisticated semi-automated pipelines, involving a powerful teacher model like GPT-4V and subsequent human refinement, to generate the necessary training data. Furthermore, innovative techniques are being explored to bridge the gap between the discrete outputs of MLLMs and the continuous nature of human quality scores, with methods like Q-Align\cite{wu2023qalignteachinglmmsvisual} and DeQA-Score\cite{you2025teachinglargelanguagemodels} focusing on modeling the distribution of human ratings rather than regressing a single value.
Recent advances have shown the effectiveness of interpretable reasoning in image quality assessment. Building on this, we extend descriptive IQA to creative image selection and propose a highly effective method.

\section{Dataset Construction}

\begin{figure*}[t]
    \centering
    \includegraphics[width=0.95\textwidth]{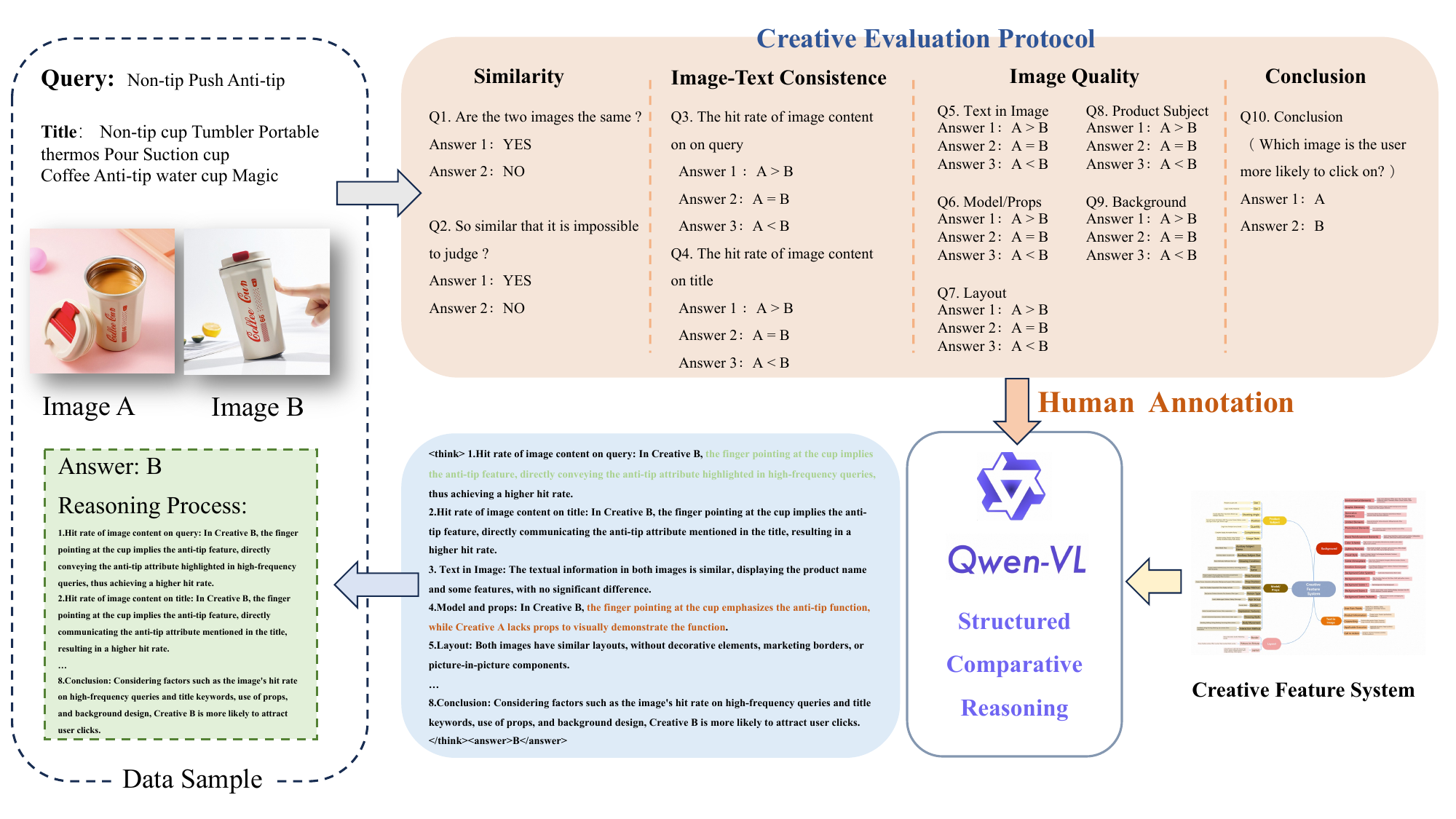}
    \vspace{-5mm}
    \caption{Annotation pipeline of CreativePair, including human annotation and comparative reasoning generation.}
    \label{fig:dataset}
\end{figure*}

High-quality datasets play a crucial role in fine-tuning multimodal large language models. To the best of our knowledge, there exists no reasoning-induced dataset for creative image selection, as prior work has not systematically modeled creative images. In this study, we introduce the first comparative, reasoning-induced creative image dataset, comprising 8,879 human-annotated image pairs. Each pair is accompanied by a label indicating the superior image and a detailed reasoning process that justifies the label. In this section, we first present our modeling approach for creative images in Sec. \ref{sec:3.1}, followed by a comprehensive description of the dataset construction process in Sec. \ref{sec:3.2}, which is also illustrated in Fig. \ref{fig:dataset}.

\subsection{Creative Image Modeling}
\label{sec:3.1}

\begin{figure}[t]
    \centering
    \includegraphics[width=\linewidth]{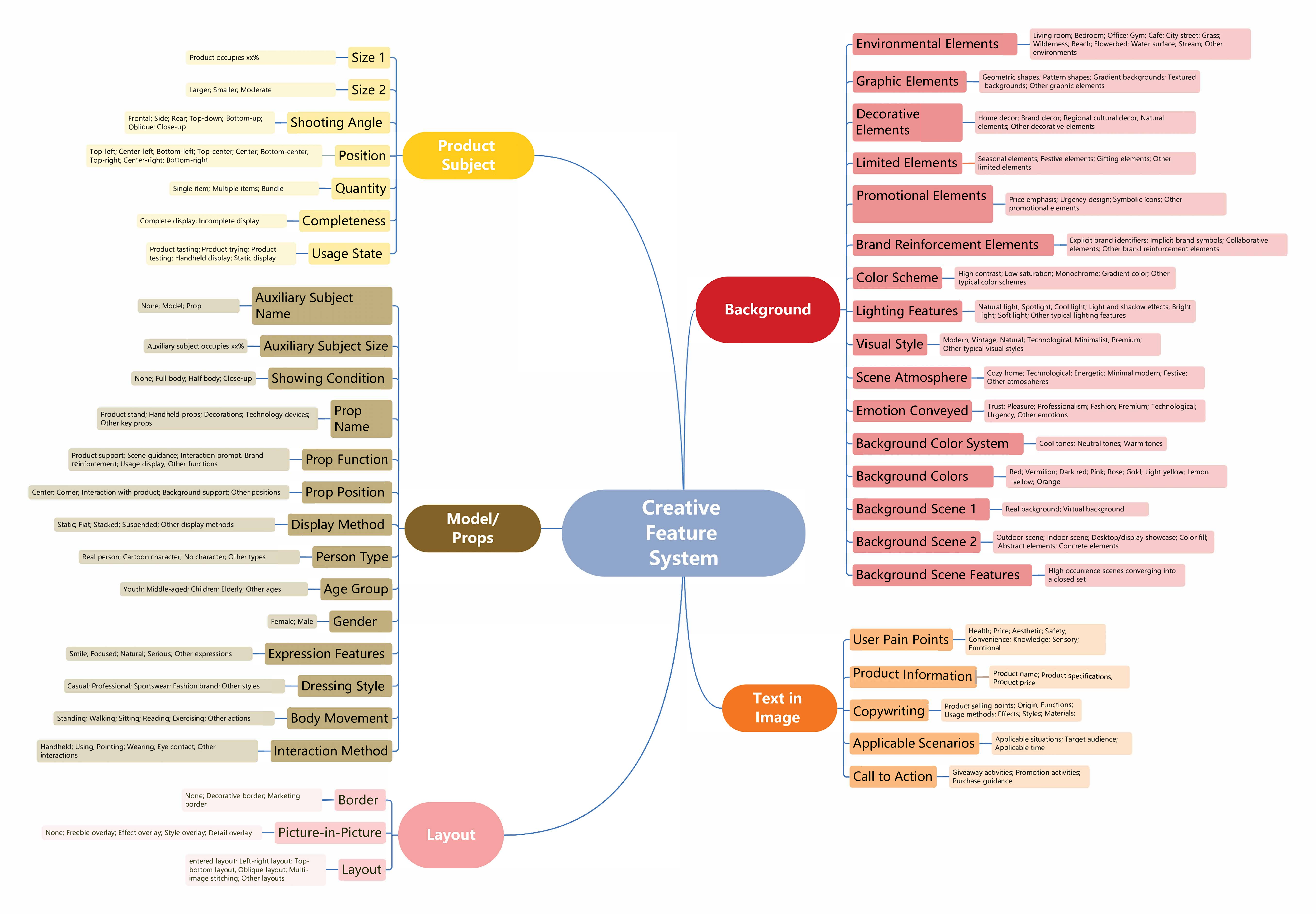}
    \caption{The overview of Creative Feature System.}
    \label{fig:Creative_Feature_System}
\end{figure}

To accurately select creative images that are more likely to attract user engagement, it is essential to first comprehensively and systematically extract informative features from the image content. In response to this need, we collaborated with professional creative image designers to develop a structured feature system, termed the Creative Feature System, as illustrated in Fig. \ref{fig:Creative_Feature_System}. 
This system categorizes the key elements of creative images into five primary components: Product Subject, Model\&Props, Background, Layout, and Text in Image. Each primary category is further decomposed into multiple subcategories, each associated with a set of attribute values. 
For instance, under the \textit{Product Subject} category, there are five subcategories: \textit{Size}, \textit{Shooting Angle}, \textit{Position}, \textit{Quantity}, and \textit{Completeness}. Among these, the \textit{Usage State} subcategory includes several attribute values such as \textit{Product Tasting}, \textit{Product Trying}, \textit{Product Testing}, \textit{Handheld Display}, and \textit{Static Display}. Notably, our proposed Creative Feature System constitutes the first systematic framework for characterizing creative images, enabling a comprehensive and structured modeling of both their visual and semantic attributes.

\subsection{Construction Pipeline}
\label{sec:3.2}

\noindent\textbf{Image pair collection.}
To model real-world user shopping scenarios, we collect creative image pairs from online traffic. 
A typical scenario on e-commerce platforms involves a user entering a target product description—referred to as \textit{query}—into the search interface. In response, the platform displays a list of relevant products, each typically accompanied by a creative image and a textual title.
To capture aspects of user intent that are most closely aligned with product characteristics, we further filter the queries, retaining only those terms identified as adjectives.

\noindent\textbf{Human annotation.}
Following the collection phase, human labeling is required for each sample. A common yet simplistic strategy involves asking annotators to visually inspect a pair of images and select their preferred option based on their first impressions. However, this method is highly subjective and lacks interpretability, offering limited justification for the final selection. 
To enhance both the explainability and consistency of the annotation process, we propose a structured evaluation protocol for creative image pairs, termed the \textbf{Creative Evaluation Protocol}, as detailed in Algorithm \ref{alg:Questions}. Specifically, our annotation process includes a pre-annotation stage preceding the formal annotation stage. In this preliminary phase, a random subset (10\%) of the collected samples is selected and annotated by professional creative image designers using the proposed Creative Feature System, along with free-form reasoning. Based on these annotations, we identify the key factors influencing the evaluation of creative images, which are then distilled into a concise and streamlined evaluation protocol.

As outlined in Algorithm \ref{alg:Questions}, the evaluation follows a four-step framework: (1) similarity assessment (Q1–Q2), (2)  Image-Text consistence evaluation (Q3–Q4), (3) image quality inspection (Q5–Q9), and (4) final decision-making (Q10). 
This structured approach not only ensures the systematic collection of rich annotation data but also significantly enhances the transparency and justifiability of the annotation decisions. After the annotation process, we filter out image pairs in which the responses to Q1 or Q2 are labeled as ``YES'', aiming to exclude pairs that are visually or semantically indistinguishable and thus not informative for model training.

\newcommand{\EmptyBox}{$\square$}

\begin{algorithm}
  \caption{Creative Evaluation Protocol}
  \label{alg:Questions}
  \begin{flushleft}

    \textbf{Q 1--2. Similarity}

    \begin{enumerate}
      \item Are the two images the \textbf{same}?\\
            \EmptyBox\ YES \quad \EmptyBox\ NO

      \item The two images are \textbf{very similar}, making it impossible to make a judgment?\\
            \EmptyBox\ YES \quad \EmptyBox\ NO
    \end{enumerate}

    \textbf{Q 3--4. Image-Text Consistence}

    \begin{enumerate}
      \setcounter{enumi}{2}
      \item The hit rate of image content on \textbf{query} (based on text in the image; elements in the image; conveyed visual style)\\
            \EmptyBox\ A > B \quad \EmptyBox\ A = B \quad \EmptyBox\ A < B

      \item The hit rate of image content on \textbf{title} (based on text in the image; elements in the image; conveyed visual style)\\
            \EmptyBox\ A > B \quad \EmptyBox\ A = B \quad \EmptyBox\ A < B
    \end{enumerate}

    \textbf{Q 5--9. Image Quality}

    \begin{enumerate}
      \setcounter{enumi}{4}
      \item \textbf{Text in Image} (whether the image contains product information text; whether it addresses user pain points, such as promises of no additives, home delivery, seven-day no-questions-asked returns, compensation for fakes, price matching guarantees, etc.; whether it highlights product selling points, such as functionality, production process, material composition, and usage effects; whether it includes calls to action, such as promotional activities, giveaway activities, and purchase guidance; whether it clearly indicates applicable scenarios)\\
            \EmptyBox\ A > B \quad \EmptyBox\ A = B \quad \EmptyBox\ A < B

      \item \textbf{Models and Props} (whether there are models/props; whether the appearance of models/props highlights the function of the product)\\
            \EmptyBox\ A > B \quad \EmptyBox\ A = B \quad \EmptyBox\ A < B

      \item \textbf{Layout} (whether the image has decorative/marketing borders; whether there is picture-in-picture, such as detail images or freebie overlays)\\
            \EmptyBox\ A > B \quad \EmptyBox\ A = B \quad \EmptyBox\ A < B

      \item \textbf{Product Subject} (ideal display effects: centered position, moderate size, suitable angle, complete subject; product quantity: multiple types/multiple colors > single type/single color; product usage state: method/state display > static display)\\
            \EmptyBox\ A > B \quad \EmptyBox\ A = B \quad \EmptyBox\ A < B

      \item \textbf{Background Design} (background color: high-end tones > solid colors, low-key colors > bright colors, ensuring the product is the visual focus; scene and atmosphere: consistent with the product > inconsistent; background image: clean and aesthetically pleasing > cluttered)\\
            \EmptyBox\ A > B \quad \EmptyBox\ A = B \quad \EmptyBox\ A < B
    \end{enumerate}

    \textbf{Q 10. Conclusion}

    \begin{enumerate}
      \setcounter{enumi}{9}
      \item Which image is the user more likely to click on?\\
            \EmptyBox\ A \quad \EmptyBox\ B
    \end{enumerate}

  \end{flushleft}
\end{algorithm}

\noindent\textbf{Chain-of-Thought generation.}
To enable Supervised Fine-Tuning (SFT) with Chain-of-Thought (CoT) reasoning, it is essential to convert the structured human annotations into coherent and natural language-based reasoning chains. To this end, we utilize \textit{Qwen2.5-VL-72B-Instruct} to generate CoT-style reasoning data for each question in the Creative Evaluation Protocol. Importantly, we enforce strict consistency between the generated reasoning and the answer in each question, ensuring that each reasoning chain logically supports and aligns with the corresponding annotation decision.
To further guarantee CoT quality, our labeling platform audits every generated reasoning chain against two criteria: (i) being non-hallucinated and (ii) faithful to the Creative Evaluation Protocol, yielding a pass rate of \textbf{95.3\%}; failed cases are iteratively refined with \textit{Qwen2.5-VL-72B-Instruct} under human supervision until both criteria are met.

\noindent\textbf{Dataset split.} After collecting the image pairs, we filter and retain 10,000 pairs based on predefined criteria, dividing them into a training set and a test set in an 8:2 ratio. Following human annotation, we remove ambiguous image pairs that are difficult to distinguish. The final training set contains 7,088 image pairs, and the test set contains 1,791 image pairs.

\section{Methodology}
\label{sec:method}

\begin{figure*}
    \centering
    \includegraphics[width=\textwidth]{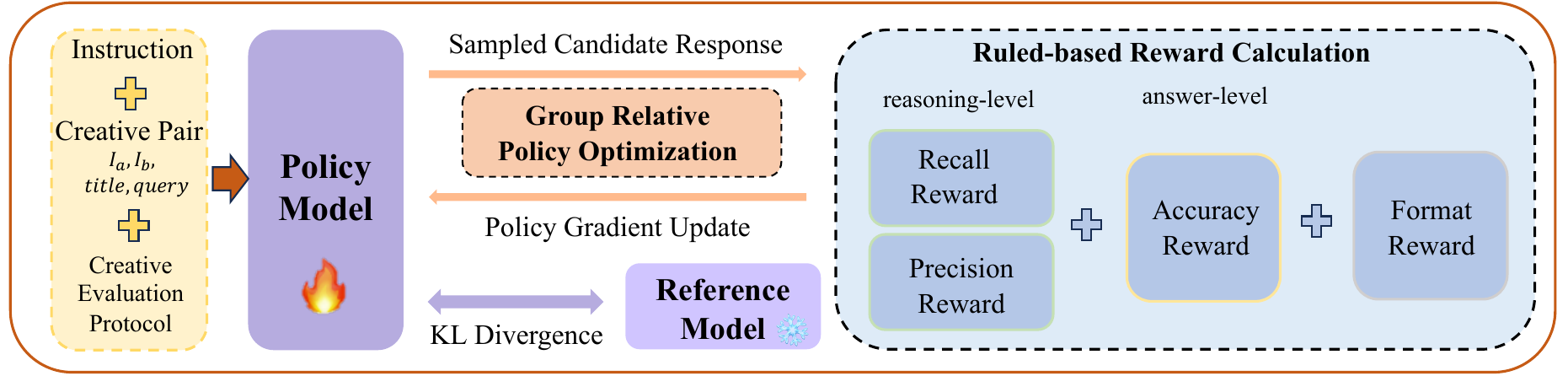}
    \caption{The overview of Group Relative Policy Optimization with Recall and Precision Reward.}
    \label{fig:pipeline}
\end{figure*}

\subsection{Problem Definition}
Explainable creative image selection can be formally defined as follows: Given two creative images for a product, denoted as $I_a$ and $ I_b $, along with the product title $title$ and a set of high-frequency user queries $ query$ associated with the product, our goal is to determine which image is more likely to be clicked by users in the context of $title$ and $query$. Moreover, a detailed comparative reasoning $cot$ must be generated before arriving at the final answer $a$. This task can be formulated as a mapping:
\begin{equation}
    M: (I_a, I_b, title, query) \xrightarrow{} (cot, a),
\end{equation}
where $I \in \mathbb{R}^{H\times W \times 3}$ is the visual input with $H$ and $W$ denoting the height and width, respectively. The product title $title$ and user query $query$ are typically expressed in natural language. Both the comparative reasoning $cot$ and the final answer $a$ are also represented as natural language outputs. Through this formulation, we can effectively leverage powerful multimodal large language models to perform explainable and accurate creative image selection.

\subsection{GRPO with Recall and Precision Reward}
\label{sec:grpo}

In this work, we choose Qwen2.5-VL-7B-Instruct as the base model for Creative4U.
To endow Creative4U with robust comparative reasoning and accurate decision-making in creative image selection, we propose Group Relative Policy Optimization with Recall and Precision Reward (GRPO-RPR), which builds on the GRPO~\citep{shao2024deepseekmath} framework. In addition to the default GRPO rewards on response format and answer correctness, GRPO-RPR introduces reasoning-level recall and precision rewards to jointly improve both the correctness of selection and the quality of explanations. Fig.~\ref{fig:pipeline} provides an overview of the proposed framework.

\noindent\textbf{Prompt Design and Output Format}. The input is designed as: 
\begin{quote}
    Please answer each question in the Creative Evaluation Protocol based on the high-frequency queries and product information, providing detailed explanations for your answers. \\
    High-frequency queries: \{query\}; \\
    Product title: \{title\}; \\
    Creative A: <image>, Creative B: <image>; \\
    \{Creative Evaluation Protocol\}; \\
    \{Output Format\}.
\end{quote}
The \textit{<image>} is the image placeholder, which will be replaced by the visual tokens of $I_a$ and $I_b$. Specifically, all outputs in the dataset follow a well-defined structure: \textit{<think>comparative reasoning process</think><answer>final answer</answer>}. As a result, the model's output during fine-tuning can be formally expressed as: \textit{y = <think>cot</think><answer>a</answer>.}

\noindent{\textbf{Sampling Response Groups.}} For each input $x$, GRPO-RPR samples a group of responses $\{y_1, y_2, \ldots, y_G\}$ from the current policy $\pi_{\theta}$. The sampling process is as follows:
\begin{equation}
y_i \sim \pi_{\theta}(y \mid x), \quad \text{for } i = 1, 2, \ldots, G.
\end{equation}
where $y_i = \{cot_i, a_i\}$. This strategy ensures diverse responses, guaranteeing sufficient and effective policy exploration.

\noindent{\textbf{Reward Calculation.}} 
Each sampled response $y_i$ is assigned a scalar reward $R(y_i)$ based on predefined and verifiable criteria, resulting in a reward set $\{r_1, r_2, \ldots, r_G\}$. In the context of the creative image selection task, the reward function $R(y_i)$ is composed of four components: a format reward $R_{format}(y_i)$, an accuracy reward $R_{acc}(y_i)$, a cot recall reward $R_{rec}(y_i)$ and cot precision reward $R_{prec}(y_i)$.  The overall reward is defined as follows:
\begin{equation}
\label{eq:reward}
R(y_i) = R_{acc}(y_i) + \alpha_1R_{format}(y_i) + \alpha_2R_{rec}(y_i) + \alpha_3R_{prec}(y_i),
\end{equation}
where $\alpha_1$,$\alpha_2$,$\alpha_3$ are weighting coefficients that balance the contributions. The format reward, $R_{format}(y_i)$, ensures that model responses are both structured and interpretable by enforcing compliance with a specified output format. Specifically, it requires the reasoning process to be enclosed within \textit{<think>} tags and the final decision to appear within \textit{<answer>} tags. A reward of 1 is granted for full compliance, while any deviation from this structure results in a zero reward. The accuracy reward, $R_{acc}(y_i)$, assesses the correctness of the model's predictions relative to the ground truth label. For the creative image selection task, it is defined as:
\begin{equation}
R_{\text{acc}}(y_i) = 
\begin{cases} 
1, & \text{if } a_{i} = a_{gt} \\ 
0, & \text{otherwise}, 
\end{cases}
\end{equation}
where $a_i$ denotes the predicted answer and $a_{gt}$ the ground truth. This binary reward penalizes deviations from the ground truth.

\textit{\textbf{Reasoning-level rewards.}} $R_{rec}(y_i)$ and $R_{prec}(y_i)$ are designed to explicitly evaluate the quality of the model’s chain-of-thought (CoT) during comparative reasoning. Each response $y_i$ contains a structured reasoning trace enclosed in \textit{<think>} tags, which we parse into a set of question--answer pairs. For a given sample, we denote the ground-truth CoT annotation and the model-generated CoT as
\begin{equation}
\begin{aligned}
&\text{CoT}_{\text{gt}}
= \bigl\{\, \langle q, a^{\text{gt}}_q \rangle \,\bigm|\, q \in \mathcal{Q}_{\text{gt}} \bigr\}\\
&\text{CoT}_{\text{pred}}
= \bigl\{\, \langle q, a^{\text{pred}}_q \rangle \,\bigm|\, q \in \mathcal{Q}_{\text{pred}} \bigr\}
\end{aligned}
\end{equation}
where each element $\langle q, a_q \rangle$ consists of a question $q$ (one of Q3--Q10 in Creative Evaluation Protocol) and its corresponding answer $a_q$. Here, each question corresponds to one dimension of comparative reasoning. We emphasize that we only retain those dimensions where the two images differ in $\mathcal{Q}_{\text{gt}}$, and we encourage the model to reason only over dimensions where it perceives a difference, in order to achieve more efficient reasoning.
We define the set of correctly answered questions as
\begin{equation}
\mathcal{Q}_{correct} =
\left\{\, q \,\middle|\,
q \in \mathcal{Q}_{\text{gt}} \cap \mathcal{Q}_{\text{pred}},
\ a^{\text{pred}}_q = a^{\text{gt}}_q
\right\}.
\end{equation}

The CoT recall and precision reward are defined as

\begin{equation}
\begin{aligned}
&R_{rec}(y_i) =
\dfrac{|\mathcal{Q}_{\text{correct}}|}{|\mathcal{Q}_{\text{gt}}|} \\
&R_{prec}(y_i) =
\dfrac{|\mathcal{Q}_{\text{correct}}|}{|\mathcal{Q}_{\text{pred}}|}
\end{aligned}
\end{equation}




By jointly maximizing both recall and precision at the reasoning level, GRPO-RPR encourages the policy to produce chains-of-thought that are both comprehensive and reliable: the model is guided to cover the key visual differences between two creatives, yet avoid noisy or spurious reasoning steps. This dual constraint leads to more effective and efficient comparative reasoning.

\noindent{\textbf{Policy Update with Relative Advantage.}} Rewards are normalized within the sampled group to compute relative advantages $\{A_1, A_2, \ldots, A_G\}$, defined as:
\begin{equation}
A_i = \frac{r_i - \text{mean}\{r_1, r_2, \ldots, r_G\}}{\text{std}\{r_1, r_2, \ldots, r_G\}}.
\end{equation}

The loss function is designed to optimize the policy by maximizing the expected advantage of the selected responses, while incorporating KL divergence as a regularization term to control the magnitude of policy updates. Ultimately, the optimization objective of GRPO can be formally expressed as follows:
\begin{multline}
\mathcal{J}_{\text{GRPO}}(\theta) = \mathbb{E}_{x, \{y_i\}} \frac{1}{G} \sum_{i=1}^{G} \frac{1}{|y_i|} \sum_{t=1}^{|y_i|} \left\{ \min\left[ \rho_{\theta/\theta_{\text{old}}}(y_{i,t}) A_{i,t}, \right. \right. \\
\left. \left. \text{clip}(\rho_{\theta/\theta_{\text{old}}}(y_{i,t}), 1 - \varepsilon, 1 + \varepsilon) A_{i,t} \right] - \beta \cdot \text{KL}(\pi_\theta \parallel \pi_{\text{ref}}) \right\},
\end{multline}
where $ \rho_{\theta/\theta_{\text{old}}}(y_{i,t}) = \frac{\pi_\theta(y_{i,t}|x)}{\pi_{\theta_{\text{old}}}(y_{i,t}|x)} $ is the ratio of probabilities under the new and old policies at step $ t $ for response $ y_i $, $ A_{i,t} $ is the normalized advantage at step $ t $ for the $ i $-th response in the group, $ \text{clip} $ is used to limit the maximum change in the probability ratio, $ \text{KL}(\pi_\theta \parallel \pi_{\text{ref}}) $ is the Kullback-Leibler divergence between the current policy $ \pi_\theta $ and a reference policy $ \pi_{\text{ref}} $, which is Qwen2.5-VL-7B-Instruct, $ \beta $ is a scalar weight controlling the strength of the KL regularization.

\section{Experiments}

\begin{figure*}[t]
    \centering
    \includegraphics[width=\textwidth]{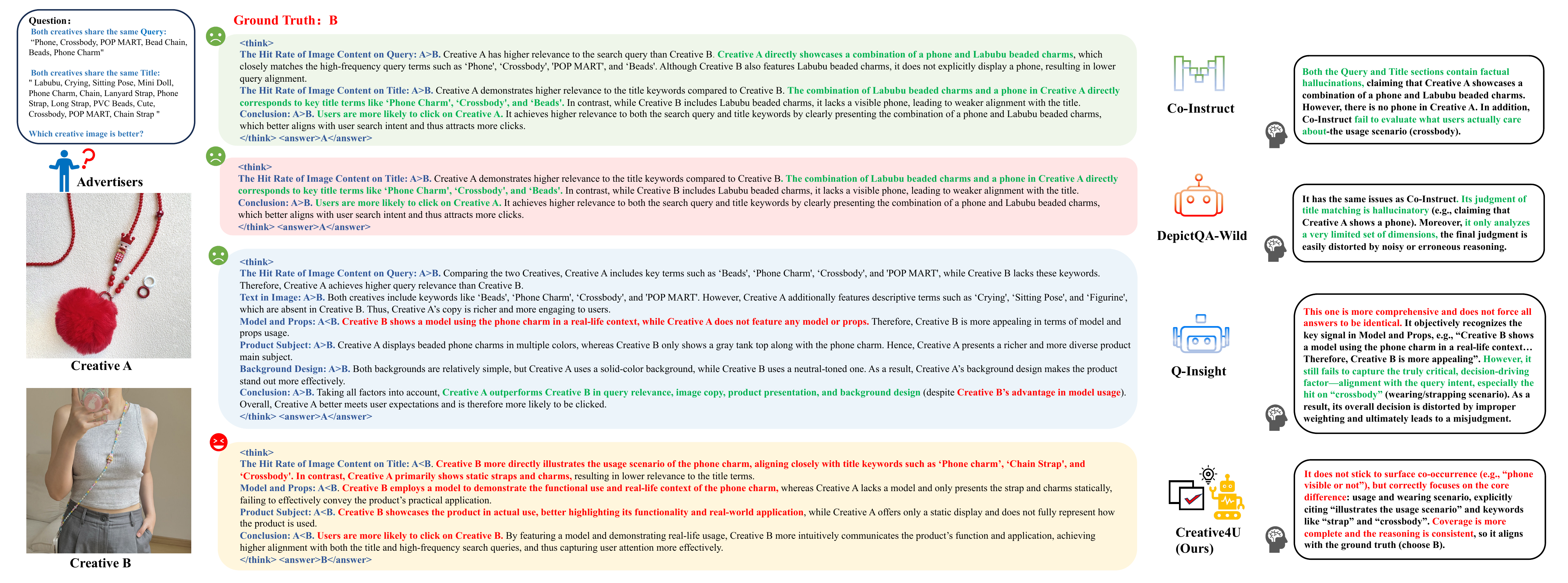}
    \caption{The qualitative comparison vs SOTA methods. Due to space constraints, we only display the key parts of the model's output. \textcolor{ForestGreen}{Green} highlights incorrect reasoning steps, while \textcolor{red}{red} emphasizes correct and critical analysis.}
    \label{fig:compare}
\end{figure*}

\subsection{Implementation details}
\textbf{Baselines and test set.} We compare with 4 open-source baselines, including VAM~\citep{wang2021hybrid}, Co-Instruct~\citep{wu2024towards},
DepictQA-Wild~\citep{you2024descriptive}, and
Q-Insight~\citep{li2025q}. VAM is a visual-aware creative ranking method and the others are comparative image quality assessment approaches. All baselines are trained on CreativePair with their default settings and evaluated on the CreativePair test set.

\noindent\textbf{Evaluation metrics.} The \textit{\textbf{accuracy}} metric is typically utilized for \textit{creative image selection} tasks. 
In this task, there are only two possible ground truth values for the final answer: A and B. 
To quantitatively evaluate the quality of comparative reasoning, we further adopt three reasoning-level metrics: \textbf{recall}, \textbf{precision} and \textbf{F1}. 
The detailed computation can be found in Sec.~\ref{sec:grpo}.
Recall, precision, and F1 are defined with respect to our Creative Evaluation Protocol and thus measure alignment with annotated comparison dimensions rather than open-ended fluency alone.
For \textit{naturalness}, the protocol is a distillation of professional marketing standards embodied in the Creative Feature System (Fig.~\ref{fig:Creative_Feature_System}), offering a comprehensive set of candidate evaluation factors rather than a rigid template; when comparing a pair, Creative4U adaptively selects the protocol dimensions it perceives as relevant (Sec.~\ref{sec:grpo}).
For \textit{usefulness}, the same recall--precision metrics intentionally target key differentiating dimensions: recall encourages covering necessary comparison aspects, while precision ensures the correctness of the stated arguments and penalizes spurious reasoning.

\noindent\textbf{Training.} Our implementation is built on the open-source
framework SWIFT\citep{zhao2024swiftascalablelightweightinfrastructure} for its scalable lightweight infrastructure for fine-tuning. In our experiment, we adopt Qwen2.5-VL-7B-Instruct\citep{bai2025qwen2} as the base model. 
In the GRPO stage, the generation number $G$ is set to 8, while reward weight $\alpha_1$,$\alpha_2$,$\alpha_3$ and KL weight $\beta$ are set to 0.1,0.5,0.5 and 0.001. 
Training is completed in approximately one day using 32 NVIDIA H20 GPUs. During inference, the average time cost per sample is 0.71 seconds.

\subsection{Qualitative comparison}

Advertisers often face the dilemma of choosing which creative image to showcase. As illustrated in Fig.~\ref{fig:compare}, given a pair of creative images, along with the high-frequency user query and the title corresponding to the products, different models conduct comparative reasoning and selections. Overall, Creative4U can make faithful comparative reasoning, leading to correct decisions. It adaptively highlights decision-critical differences between creatives rather than mechanically enumerating every protocol dimension, yielding rationales that read as natural rather than template-like. In contrast, other models are hindered by hallucination during their reasoning process, ultimately failing to make accurate selections.

Using the phone charm case in Fig.~\ref{fig:compare} as an example, Creative4U accurately identifies the key decision-driving factor: the alignment between the image and the crossbody usage scenario implied by the query and title. While other models are misled by superficial signals (e.g., whether a phone is explicitly visible) and even hallucinate that Creative A “shows a phone”, Creative4U focuses on whether the wearing/strapping scenario is clearly demonstrated. It points out that Creative B, by featuring a model actually using the phone charm in a crossbody context, better matches the intent behind keywords such as “phone charm”, “chain strap”, and “crossbody”. This reasoning leads Creative4U to correctly select Creative B, whereas Co-Instruct, DepictQA-Wild, and Q-Insight all make incorrect selections driven by noisy or erroneous comparative reasoning.


\begin{table}[h]
    \centering
    \caption{Quantitative comparisons vs SOTA methods.}
    \vspace{-2mm}
    \begin{tabular}{@{}lcccc@{}}
        \toprule
        Method & accuracy$\uparrow$ & recall$\uparrow$ & precision$\uparrow$ & F1$\uparrow$ \\ \midrule
        VAM~\citep{wang2021hybrid} & 0.608 & -      & -      & -      \\
        Co-Instruct~\citep{wu2024towards} & 0.515 & 0.365 & 0.365 & 0.344 \\
        DepictQA-Wild~\citep{you2024descriptive} & 0.577 & 0.405 & \textbf{0.434} & 0.399 \\
        Q-Insight~\citep{li2025q} & 0.751 & 0.255 & 0.203 & 0.213 \\
        Creative4U(ours) & \textbf{0.791} & \textbf{0.799} & 0.397 & \textbf{0.514} \\
        \bottomrule
    \end{tabular}
    \label{tab:methods}
\end{table}

\subsection{Quantitative comparison}

The quantitative results are summarized in Table~\ref{tab:methods}.
In terms of accuracy, Creative4U achieves the best performance with a score of 0.791, surpassing all competitor methods. This indicates that Creative4U more reliably selects the better creative image.
Beyond final decisions, Creative4U achieves a recall of 0.799 and a F1 of 0.514, both substantially higher than all baselines, which shows that Creative4U covers a much larger portion of the ground-truth comparison dimensions while maintaining a better overall balance between recall and precision. Although DepictQA-Wild attains the highest precision (0.434), its recall and F1  are clearly lower than those of Creative4U. Q-Insight, while strong in accuracy, exhibits relatively low reasoning-level scores, indicating that it often makes correct choices but with sparse or incomplete comparative reasoning.
Overall, these quantitative results underscore Creative4U's advantage for explainable creative image selection.

\subsection{Ablation Study}

\noindent\textbf{Training and Reward Design.}
We ablate the training paradigm and reward objectives of Creative4U, as summarized in Table~\ref{tab:ablation}.
\textit{Zero-Shot} denotes direct inference without task-specific training, \textit{CoT-SFT} is supervised fine-tuning with chain-of-thought outputs, and \textit{GRPO} is GRPO with only accuracy and format rewards, which coincides with the Q-Insight~\cite{li2025q} setting.
Compared to Zero-Shot and GRPO, CoT-SFT\cite{wei2022chain, ouyang2022training} substantially improves both accuracy and reasoning metrics, indicating the benefit of supervised comparative reasoning.
We then analyze different RL objectives.
\textit{GRPO-F1R} augments GRPO with an F1-based reasoning reward, while \textit{DAPO-RPR} and \textit{GSPO-RPR} replace GRPO with DAPO\cite{yu2025dapo} and GSPO\cite{zheng2025group} schemes using recall--precision rewards.
These variants yield mixed trade-offs between accuracy and reasoning quality, but none consistently dominates.
Finally, our proposed \textit{GRPO-RPR} achieves the best accuracy and the highest recall, while \textit{CoT-SFT + GRPO-RPR} attains the best precision and F1.
This shows that integrating recall--precision reasoning rewards into GRPO effectively enhances both decision accuracy and the coverage-precision balance of comparative reasoning.

\begin{table}[t]
    \centering
    \caption{Ablation analysis of training and reward designs.}
    \vspace{-2mm}
    \begin{tabular}{@{}lcccc@{}}
        \toprule
        Method & accuracy$\uparrow$ & recall$\uparrow$ & precision$\uparrow$ & F1$\uparrow$ \\ \midrule
        Zero-Shot & 0.577 & 0.182 & 0.163 & 0.159 \\
        CoT-SFT & 0.767 & 0.652 & 0.667 & 0.642 \\
        GRPO & 0.751 & 0.255 & 0.203 & 0.213 \\
        \midrule
        GRPO-F1R & 0.767 & 0.692 & 0.513 & 0.569 \\
        DAPO-RPR & 0.762 & 0.371 & 0.476 & 0.406 \\
        GSPO-RPR & 0.758 & 0.773 & 0.386 & 0.498 \\
        \midrule
        CoT-SFT + GRPO-RPR & 0.772 & 0.632 & \textbf{0.695} & \textbf{0.646} \\
        GRPO-RPR(ours) & \textbf{0.791} & \textbf{0.799} & 0.397 & 0.514 \\
        \bottomrule
    \end{tabular}
    \label{tab:ablation}
\end{table}

\noindent\textbf{Sensitivity of Reward Coefficients.}
We study the sensitivity of the three reward weights in Eq.~\eqref{eq:reward} by varying $(\alpha_1,\alpha_2,\alpha_3)$, as summarized in Table~\ref{tab:sensitivity}.
The format weight $\alpha_1$ is non-critical: scaling it from 0.0 to 0.5 shifts accuracy and F1 by only $\sim$1 point.
Recall and precision rewards are complementary and both indispensable: removing recall collapses F1 from 0.514 to 0.376, since the model is no longer encouraged to cover the key comparison dimensions.
Removing precision drags accuracy to the lowest 0.729, since spurious reasoning is left unchecked and misleads the final decision.
A balanced weighting yields the best trade-off: skewing toward recall maximizes recall but loses accuracy, and skewing toward precision peaks accuracy but underperforms in F1.
$\alpha_2{=}\alpha_3{=}0.5$ attains the best F1 with near-best accuracy, justifying our default setting.

\begin{table}[t]
    \centering
    \caption{Sensitivity analysis of reward coefficients $(\alpha_1,\alpha_2,\alpha_3)$ in Eq.~\eqref{eq:reward}.}
    \vspace{-2mm}
    \begin{tabular}{@{}clcccc@{}}
        \toprule
        ID & $(\alpha_1,\alpha_2,\alpha_3)$ & accuracy$\uparrow$ & recall$\uparrow$ & precision$\uparrow$ & F1$\uparrow$ \\ \midrule
        1 & (0.0, 0.5, 0.5) & 0.766 & 0.765 & 0.360 & 0.478 \\
        2 & (0.5, 0.5, 0.5) & 0.778 & 0.779 & 0.376 & 0.492 \\
        3 & (0.1, 0.0, 0.5) & 0.761 & 0.527 & 0.313 & 0.376 \\
        4 & (0.1, 0.5, 0.0) & 0.729 & 0.785 & 0.325 & 0.434 \\
        5 & (0.1, 0.7, 0.3) & 0.754 & \textbf{0.818} & 0.354 & 0.489 \\
        6 & (0.1, 0.3, 0.7) & \textbf{0.796} & 0.765 & \textbf{0.399} & 0.510 \\
        7(ours) & (0.1, 0.5, 0.5) & 0.791 & 0.799 & 0.397 & \textbf{0.514} \\
        \bottomrule
    \end{tabular}
    \label{tab:sensitivity}
\end{table}

\subsection{Deployment}
\label{sec:deployment}
Creative4U is deployed as an \textit{offline} creative selector, so it is decoupled from strict online latency budgets. In e-commerce advertising, each product already displays a current best creative $p$ and is supplied with $N$ AI generated candidates to be evaluated against $p$. Since naive enumeration over all pairs scales as $O(N^2)$, we adopt a \textbf{two-stage coarse-to-fine} pipeline. \textit{Stage 1 (Coarse Filtering)} compares each of the $N$ candidates against $p$ in one pair-wise pass and keeps only the $N'$ candidates that beat $p$, with cost $O(N)$. \textit{Stage 2 (Fine Ranking)} performs $\binom{N'}{2}$ pair-wise comparisons among the $N'$ survivors and aggregates the per-creative win frequency to produce the final ranking, with cost $O(N'^{2})$. In Taobao's advertising business, $N$ is at most 100 and the empirical filtering ratio is $N'/N \approx 12\%$, so Stage~1 shrinks the candidate set to a much smaller $N'$ and makes the Stage~2 comparisons affordable.

\subsection{Online A/B test}
In practice, creative images first go through offline selection and are then refined by an online ranker before being displayed to users. To evaluate the online business impact of the creatives selected by Creative4U, we replace the offline selector with two baselines while keeping the rest of the production pipeline intact: (1) Random, which randomly selects creatives, and (2) Bandit, following the approach used in HBM\citep{wang2021hybrid}. All A/B tests are conducted on Taobao, one of the largest e-commerce platforms in the world.

\noindent\textbf{Data \& Metrics}. We employ real-world online A/B testing to further validate the proposed approach. For each method, we predict the top-10 most effective creatives within an ad set. For Creative4U, the top-10 creatives are produced by the two-stage offline pipeline described in Sec.~\ref{sec:deployment}.
These selected creatives are then passed to the online ranking system and subsequently served to users. Each A/B test runs continuously for one month to ensure stable and reliable measurements.
We evaluate online performance using the following key metrics: Click-Through Rate (CTR), Conversion Rate (CVR), and Revenue Per Mille (RPM). These are computed as standard industry metrics. All percentage improvements are calculated relative to the \textit{Random Pick} baseline. For example, the relative improvement in CTR is computed as $\frac{CTR_1-CTR_2}{CTR_2}$. 


From Table~\ref{tab:online_evaluations}, we can observe that: (1) The Random Pick baseline exhibits the lowest performance, reinforcing the notion that intelligent creative selection significantly improves user engagement and platform revenue. (2) Creative4U consistently outperforms Bandit in both experimental settings, regardless of access to the online ranker. Without the online ranker, Creative4U achieves relative improvements of 2.4\% in CTR, 2.08\% in RPM, and 0.96\% in CVR. With the online ranker, it continues to deliver consistent gains of 1.32\% in CTR, 1.34\% in RPM, and 0.35\% in CVR, demonstrating its robustness and seamless compatibility within complex production ranking systems.

\begin{table}[t]
    \centering
    \caption{Online evaluations.}
    \vspace{-2mm}
    \begin{tabular}{@{}llcccc@{}}
        \toprule
        Online Rank & Offline Rank       & CTR(\%) $\uparrow$    & RPM(\%)  $\uparrow$   & CVR(\%)  $\uparrow$   \\ \midrule
        \multirow{3}{*}{\centering no}  & Random         & -       & -       & -       \\
                             & Bandit                & +4.67 & +4.84 & +1.52 \\
                             & Creative4U         & \textbf{+7.07} & \textbf{+6.92} & \textbf{+2.48} \\ \midrule
        \multirow{3}{*}{\centering yes} & Random         & -       & -       & -       \\
                             & Bandit                & +2.43 & +2.52 & +1.34 \\
                             & Creative4U         & \textbf{+3.75} & \textbf{+3.86} & \textbf{+1.69} \\ \bottomrule
    \end{tabular}
    \label{tab:online_evaluations}
\end{table}

\subsection{Explainability in Practice}
Beyond final-decision accuracy, the comparative rationales generated by Creative4U are highly valuable in practical creative image production. Specifically, our AIGC team leverages Creative4U's dimension-level feedback to conduct multiple iterations of creative generation, including (i) \textit{background optimization}, generating backgrounds tailored to the semantics of query words and the actual usage scenarios; and (ii) \textit{text-in-image refinement}, refining text overlaid on the image by combining query intent with product selling points. Through such feedback-driven iterations, the win rate of AIGC creatives against human-made ones improved from \textbf{12.2\% to 44.5\%}, demonstrating Creative4U's practical value as actionable guidance for designers.
\section{Conclusion and Discussion}
In this paper, we introduce Creative4U, a pioneering framework for explainable assessment and selection of creative images in e-commerce advertising. Our work addresses the critical gap in existing methodologies, enhancing both the efficiency of image selection and providing clear reasoning for these decisions.
The development of the CreativePair dataset marks a significant advancement in comparative reasoning for creative image, facilitating a systematic evaluation based on detailed annotations and user intent. 
Our comprehensive offline and online experiments demonstrate that Creative4U, utilizing multimodal large language models, significantly outperforms existing methods in selecting high-performing creative images. By integrating real user queries into the selection process, we ensure alignment with user preferences, ultimately improving advertising effectiveness and user experience.

In this study, we validate the effectiveness of explainable creative image selection through online experiments. This encouraging result has sparked our imagination about future directions, such as converting the reasoning process into quantifiable scores used for providing feedback on AI-generated images, thereby promoting the generation of higher-quality visuals.

\clearpage
\bibliographystyle{ACM-Reference-Format}
\balance
\bibliography{acmart}

\end{document}